\documentclass[11pt,a4paper]{article}
\usepackage[hyperref]{emnlp-ijcnlp-2019}
\usepackage{times}
\usepackage{latexsym}
 \usepackage{graphicx}
\usepackage[normalem]{ulem}
\usepackage{subfigure}
\usepackage{url}

\aclfinalcopy 

\title{Discrete Argument Representation Learning for Interactive Argument Pair Identification}

\author { Lu Ji$^1$ , 
Zhongyu Wei$^{2}$\thanks{*Corresponding author} , 
Jing Li$^3$, 
Qi Zhang$^1$ , 
Xuanjing Huang$^1$\\
$^1$School of Computer Science, Fudan University, Shanghai, China 
\\ $^2$School of Data Science, Fudan University, China
\\ $^3$Hong Kong Polytechnic University \\
{\tt  \{17210240034,zywei,qi\_zhang,xjhuang\}@fudan.edu.cn } \\
{\tt lijing.nlp@link.cuhk.edu.hk}
}

\date{}

\begin{document}
\maketitle
\begin{abstract}
  In this paper, we focus on extracting interactive argument pairs from two posts with opposite stances to a certain topic. Considering opinions are exchanged from different perspectives of the discussing topic, we study the discrete representations for arguments to capture varying aspects in argumentation languages (e.g., the debate focus and the participant behavior). Moreover, we utilize hierarchical structure to model post-wise information incorporating contextual knowledge. Experimental results on the large-scale dataset collected from \emph{CMV} show that our proposed framework can significantly outperform the competitive baselines. Further analyses reveal why our model yields superior performance and prove the usefulness of our learned representations.
\end{abstract}
\section{Introduction}
Arguments play a central role in decision making on social issues. 
Striving to automatically understand human arguments, computational argumentation becomes a growing field in natural language processing. 
It can be analyzed at two levels --- monological argumentation and dialogical argumentation. Existing research on monological argumentation covers argument structure prediction~\cite{Stab:14}, claims generation~\cite{Bilu:16}, essay scoring~\cite{Taghipour:16}, etc. 
Recently, dialogical argumentation becomes an active topic.

\begin{figure}[t!]
\setlength{\belowcaptionskip}{-0.7cm} 
\small 
\begin{tabular}{p{0.455\textwidth}}
\hline 
\multicolumn{1}{|p{0.455 \textwidth}|}{\textit{CMV: The position of vice president of the USA should be eliminated from our government.}} \\
\hline
\multicolumn{1}{|p{0.455 \textwidth}|}{\textbf{Post A:}\ \  \textbf{a1:} \dotuline{[If the president is either killed or resigns, the vice president is a horrible choice to take over office.]} \textbf{a2:} The speaker of the House would be more qualified for the position. \textbf{a3:} \uwave{[I'm willing to bet that John Boehner would have an easier time dealing with congress as president than Joe Biden would due to his constant interaction with it.]} \textbf{a4:} If Boehner took office, as a republican, would he do something to veto bills Obama supported?} \\
\hline 
\multicolumn{1}{|p{0.455 \textwidth}|}{\textbf{Post B:}\ \ \textbf{b1:}\dotuline{[Seriously, stop this hyperbole.]} \textbf{b2:} \uwave{[Do you think that have anything to do with the fact that Boehner is a republican, and republicans control congress?]} \textbf{b3:} That argument has much less to do with the individuals than it does with the current party in control.}\\
\hline
\end{tabular}
\caption{\label{font-table} An example of dialogical argumentation consists of two posts from change my view, a sub-forum of Reddit.com. Different types of underlines are used to highlight the interactive argument pairs.
}
\label{figure0}
\end{figure}
In the process of dialogical arguments, participants exchange arguments on a given topic~\cite{Asterhan:07, Hunter:13}. 
With the popularity of online debating forums, large volume of dialogical arguments are daily formed, concerning wide rage of topics. 
A social media dialogical argumentation example from ChangeMyView subreddit is shown in Figure~\ref{figure0}.  
There we show two posts holding opposite stances over the same topic. 
One is the original post and the other is reply. 
As can be seen, opinions from both sides are voiced with multiple arguments and the reply \emph{post B} is organized in-line with \emph{post A}'s arguments. 
Here we define an interactive argument pair formed with two arguments from both sides (with the same underline), which focuses on the same perspective of the discussion topic.
The automatic identification of these pairs will be a fundamental step towards the understanding of dialogical argumentative structure. Moreover, it can benefit downstream tasks, such as debate summarization~\cite{Sanchan:17} and logical chain extraction in debates~\cite{Botschen:18}.

However, it is non-trivial to extract the interactive argument pairs holding opposite stances.
Back to the example. Given the argument \textbf{b1} with only four words contained, it is difficult, without richer contextual information, to understand why it has interactive relationship with \textbf{a1}. 
Therefore, without modeling the debating focuses of arguments, it is likely for models to wrongly predict
interactive relationship with \textbf{a4} for sharing more words. Motivated by these observations, we propose to explore discrete argument representations to capture varying aspects (e.g., the debate focus) in argumentation language and learn context-sensitive argumentative representations for the automatic identification of interactive argument pairs. 



For argument representation learning, different from previous methods focusing on the modeling of continuous argument representations, we obtain discrete latent representations via discrete variational autoencoders and investigate their effects on the understanding of dialogical argumentative structure. For context representation modeling, we employ a hierarchical neural network to explore what content an argument conveys and how they interact with each other in the argumentative structure. 
To the best of our knowledge, we are the first to explore discrete representations on argumentative structure understanding. 
In model evaluation, we construct a dataset collected from \emph{CMV}~\footnote{\url{https://reddit.com/r/changemyview}}, which is built as part of our work and has be released~\footnote{\url{http://www.sdspeople.fudan.edu.cn/zywei/data/arg-pairs-fudanU.zip}}. 
Experimental results show that our proposed model can significantly outperform the competitive baselines. 
Further analysis on discrete latent variables reveals why our model yields superior performance. 
At last, we show that the  representations learned by our model can successfully boost the performance of argument persuasiveness evaluation.


\section{Task Definition and Dataset Collection}
In this section, we first define our task of interactive argument pair identification, followed by a description of how we collect the data for this task.

\subsection{Task Definition} 
Given a argument $q$ from the original post, a candidate set of replies consisting of one positive reply $r^+$, several negative replies $r_1^-\sim r_u^-$, and their corresponding argumentative contexts, our goal is to automatically identify which reply has interactive relationship with the quotation $q$.

\begin{table}[t!]\small
\setlength{\belowcaptionskip}{-0.7cm} 
\begin{center}
\begin{tabular}{l|c|c}
\hline & training set &test set \\ \hline
\# of arg. per post & 11.8$\pm$6.6 & 11.4$\pm$6.2\\
\# of token per post  & 209.7$\pm$117.2 & 205.9$\pm$114.6\\\hline
\# of token per $q$  & 20.0$\pm$8.6 & 20.0$\pm$8.6 \\
\# of token per $p_r$ &16.9$\pm$8.1 &17.3$\pm$8.4\\
\# of token per $n_r$ &19.0$\pm$8.0 &19.1$\pm$8.1 \\ \hline
max \# of $q$-$p_r$ pairs  &12 &9\\ 
avg. \# of $q$-$p_r$ pairs  &1.5$\pm$0.9 &1.4$\pm$0.9 \\\hline
\end{tabular}
\caption{\label{font-table} Overview statistics of the constructed dataset (mean and standard deviation). \emph{arg.}, $q$, $p_r$, $n_r$ represent \emph{argument}, \emph{quotation}, \emph{positive reply} and \emph{negative reply} respectively. $q$-$p_r$ represents the quotation-reply pair between posts.}
\label{tab:tb1}
\end{center}
\end{table}
We formulate the task of identifying interactive argument pairs as a pairwise ranking problem. In practice, we calculate the matching score $S(q,r)$ for each reply in the candidate set with the quotation $q$ and treat the one with the highest matching score as the winner.

\subsection{Dataset Collection}

Our data collection is built on the \emph{CMV} dataset released by~\citeauthor{Tan:16}~\shortcite{Tan:16}.
In \emph{CMV}, users submit posts to elaborate their perspectives on a specific topic and other users are invited to argue for the other side to change the posters' stances. The original dataset is crawled using Reddit API. Discussion threads from the period between January 2013 and May 2015 are collected as training set, besides, threads between May 2015 and September 2015 are considered as test set. In total, there are 18,363 and 2,263 discussion threads in training set and test set, respectively. 



An observation on \emph{CMV} shows that when users reply to a certain argument in the original post, they quote the argument first and write responsive argument directly, forming a quotation-reply pair. Figure~\ref{figure00} shows how quotation-reply pairs could be identified.
\begin{figure}[htbp]
\setlength{\belowcaptionskip}{-0.3cm} 
\small 
\begin{tabular}{|p{0.45\textwidth}|}
\hline 
\multicolumn{1}{p{0.45 \textwidth}}{\textbf{Original Post:}\ \  ... Strong family values in society lead to great results. \textit{I want society to take positive aspects of the early Americans and implement that into society.} This would be a huge improvement than what we have now. ...}\\
\multicolumn{1}{p{0.45 \textwidth}}{\textbf{User Post:}\ \ \textit{\&gt; I want society to take positive aspects of the early Americans and implement that into society.} What do you believe those aspects to be? ...}\\
\hline
\end{tabular}
\caption{\label{font-table} An example illustrating the formation process of a quotation-reply pair in \emph{CMV}.
}
\label{figure00}
\end{figure}
\begin{figure*}[t!]
\setlength{\abovecaptionskip}{-0.03cm} 
\setlength{\belowcaptionskip}{-0.3cm} 
\centering
   \includegraphics[width=1.8 \columnwidth, height=9cm ]{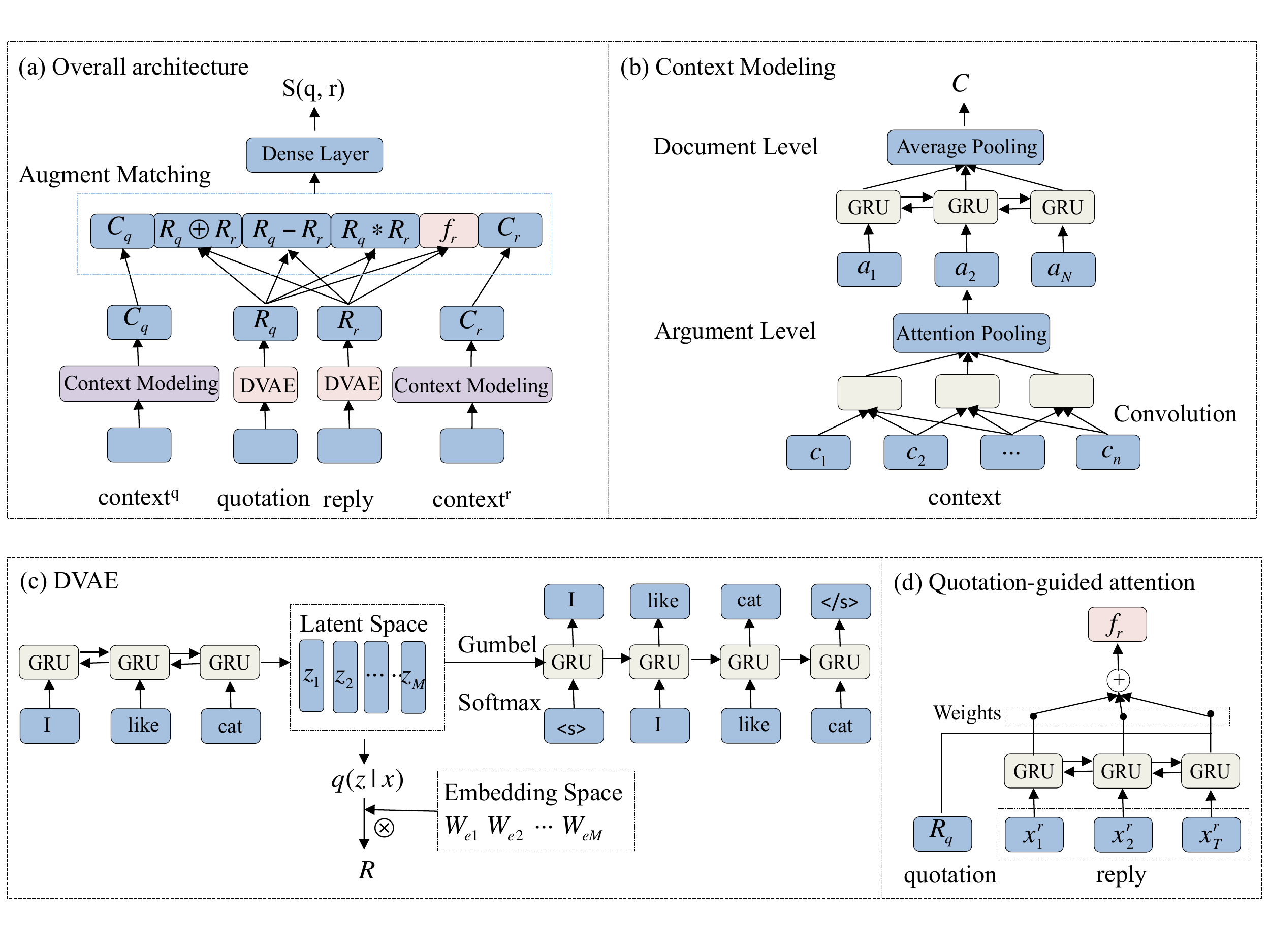}
    \caption{(a) Overall architecture of the proposed model. (b) Hierarchical architecture for argumentative context modeling. (c) Detailed structure of the discrete variational autoencoders (\emph{DVAE}). (d) Structure of the quotation-guided attention in argument matching.}
    \label{figure2}
\end{figure*}

Inspired by this finding, we decide to extract interactive argument pairs with the relation of quotation-reply. In general, the content of posts in \emph{CMV} is informal, making it difficult to parse an argument in a finer-grain with premise, conclusion and other components. Therefore, following previous setting  in~\citeauthor{Ji:2018}~\shortcite{Ji:2018}, we treat each sentence as an argument.
Specifically, we only consider the quotation containing one argument and view the first sentence after the quotation as the reply. We treat the quotation-reply pairs extracted as positive samples and randomly select four replies from other posts that are also related to the original post to pair with the quotation as negative samples. In detail, each instance in our dataset includes the quotation, one positive reply, four negative replies, and the posts where they exist. The posts where they exist refer to argumentative contexts mentioned below. What's more, we remove quotations from argumentative contexts of replies.



We keep words with the frequency higher than 15 and this makes the word vocabulary with 20,692 distinct entries. In order to assure the quality of quotation-reply pairs, we only keep the instance where the number of words in the quotation and replies range from 7 to 45. We regard the instances extracted from training set and test set in~\citeauthor{Tan:16}~\shortcite{Tan:16} for training and test. The number of instances in training and test set is 11,565 and 1,481, respectively. We randomly select 10\% of the training instances to form the development set. The statistic information of our dataset is shown in Table~\ref{tab:tb1}.


\section{Proposed Model}
The overall architecture of our model is shown in Figure~\ref{figure2}(a). It takes a quotation, a reply and their corresponding argumentative contexts as inputs, and outputs a real value as its matching score. It mainly consists of three components, namely, \textit{Discrete Variational AutoEncoders} (\emph{DVAE}, Figure~\ref{figure2}(c)), \textit{Argumentative Context Modeling} (Figure~\ref{figure2}(b)) and \textit{Argument Matching and Scoring}. We learn discrete argument representations via \emph{DVAE} and employ a hierarchical architecture to obtain the argumentative context representations. The \textit{Argument Matching and Scoring} integrates some semantic features between the quotation and the reply to calculate the matching score.

\subsection{Discrete Variational AutoEncoders}
We employ discrete variational autoencoders~\cite{Rolfe:17} to reconstruct arguments from auto-encoding and obtain argument representations based on discrete latent variables to capture different aspects of argumentation languages.

\noindent \textbf{Encoder.} Given an argument $x$ with words $w_1, w_2,..., w_T$, we first embed each word to a dense vector obtaining $w^{'}_1, w^{'}_2,..., w^{'}_T$ correspondingly. Then we use a bi-directional GRU~\cite{Wang:18} to encode the argument. 
\begin{equation}
\setlength{\abovedisplayskip}{3pt}
\setlength{\belowdisplayskip}{3pt}
{h_t} =BiGRU(w^{'}_t,h_{t-1})
\end{equation}
We obtain the hidden state for a given word $w^{'}_t$ by concatenating the forward hidden state and backward hidden state. Finally, we consider the last hidden state $h_T$ as the continuous representation of the argument.
 
\noindent \textbf{Discrete Latent Variables.} We introduce $z$ as a set of K-way categorical variables $z=\left\{{z_1,z_2,..., z_M}\right\}$, where $M$ is the number of variables. Here, each $z_i$ is independent and we can easily extend the calculation process below to every latent variables.
Firstly, we calculate the logits $l_i$ as follows.
\begin{equation}
\setlength{\abovedisplayskip}{3pt}
\setlength{\belowdisplayskip}{3pt}
{l_i} =W_{l}h_T+b_l
\end{equation}
where $W_l\in R^{{K}\times{E}}$ stands for the weight matrix, $E$ is the dimension of hidden units in encoder, while $b_l$ is a weight vector.

After obtaining the logits $l_i$, we can calculate the posterior distribution and discrete code of $z_i$.
\begin{equation}
\setlength{\abovedisplayskip}{3pt}
\setlength{\belowdisplayskip}{3pt}
{q(z_i|x)} =Softmax(l_i)
\end{equation}
\begin{equation}
\setlength{\abovedisplayskip}{3pt}
\setlength{\belowdisplayskip}{3pt}
{Z_{code}(i)}=\mathop{\arg\max}_{k \in [1,2,...,K]}(l_{ik})
\end{equation}

However, using discrete latent variables is challenging when training models end-to-end. To alleviate this problem, we use the recently proposed Gumbel-Softmax trick~\cite{Lu:17} to create a differentiable estimator for categorical variables. During training we draw samples $g_1,g_2,...,g_K$ from the Gumbel distribution: $g_k\sim -− \log(-− \log(u))$, where $u\sim U(0,1)$ are uniform samples. Then, we compute the log-softmax of $l_i$ to get $\omega_i\in R^K$:
\begin{equation}
\setlength{\abovedisplayskip}{3pt}
\setlength{\belowdisplayskip}{3pt}
{\omega_{ik}}=\frac{exp((l_{ik}+g_k)/\tau)}{\sum_{k}{exp((l_{ik}+g_k)/\tau)}}
\end{equation}

 $\tau$ is a hyper-parameter. With low temperature $\tau$, this vector $\omega_i$ is close to the one-hot vector representing the maximum index of $l_i$. But with higher temperature, this vector $\omega_i$ is smoother. 

Then we map the latent samples to the initial state of the decoder as follows: 
\begin{equation}
\setlength{\abovedisplayskip}{3pt}
\setlength{\belowdisplayskip}{3pt}
{h_{dec}^0}={\sum_{i=1}^M{W_{ei}\omega_i}}
\end{equation}
where $W_{ei}\in R^{{K}\times{D}}$ is the embedding matrix, $D$ is the dimension of hidden units in decoder. Finally, we use a GRU as the decoder to reconstruct the argument given ${h_{dec}^0}$.

\noindent \textbf{Discrete Argument Representations.} Through the process of auto-encoding mentioned above, we can reconstruct the argument. The representation that we want to find can capture varying aspects in argumentation languages and contain salient features of the argument.
$q(z_i|x)$ shows the probability distribution of $z_i$ over $K$ categories, which contains salient features of the argument on varying aspects. Therefore, we obtain the discrete argument representation by the posterior distribution of discrete latent variables $z$. 
\begin{equation}
\setlength{\abovedisplayskip}{3pt}
\setlength{\belowdisplayskip}{3pt}
{R={\sum_{i=1}^M{W_{ei}q(z_i|x)}}}
\end{equation}

\subsection{Argumentative Context Modeling}
Here, we introduce contextual information of the quotation and the reply to help identity the interactive argument pairs. The argumentative context contains a list of arguments. Following previous setting in~\citeauthor{Ji:2018}~\shortcite{Ji:2018}, we consider each sentence as an argument in the context. Inspired by~\citeauthor{Dong:17}~\shortcite{Dong:17}, we employ a hierarchical architecture to obtain argumentative context representations. 

\noindent \textbf{Argument-level CNN.} Given an argument and their embedding forms $\left\{e_1, e_2,..., e_n\right\}$, 
we employ a convolution layer to incorporate the context information on word level. 
\begin{equation}
\setlength{\abovedisplayskip}{3pt}
\setlength{\belowdisplayskip}{3pt}
{s_i} = f({W_s} \cdot [e_i:e_{i + {ws} - 1}] + {b_s})
\end{equation}
where $W_s$ and $b_s$ are weight matrix and bias vector. $ws$ is the window size in the convolution layer and $s_i$ is the feature representation. Then, we conduct an attention pooling operation over all the words to get argument embedding vectors.
\begin{equation}
\setlength{\abovedisplayskip}{5pt}
\setlength{\belowdisplayskip}{5pt}
{m_i} = {\rm{tan}}h ({W_m} \cdot {s_i} + {b_m})
\end{equation}
\begin{equation}
\setlength{\abovedisplayskip}{5pt}
\setlength{\belowdisplayskip}{5pt}
{u_i} = \frac{{{e^{{W_u \cdot {m_i}}}}}}{{\sum\limits_{j}{{e^{{W_u \cdot {m_j}}}}} }}
\end{equation}
\begin{equation}
\setlength{\abovedisplayskip}{5pt}
\setlength{\belowdisplayskip}{5pt}
a = \sum\limits_{i} {{u_i} \cdot {s_i}}
\end{equation}
where $W_m$ and $W_u$ are weight matrix and vector, $b_m$ is the bias vector, $m_i$ and $u_i$ are attention vector and attention weight of the  $i$-th word. $a$ is the argument representation.

\noindent \textbf{Document-level BiGRU.} Given the argument embedding $\left\{a_1, a_2,..., a_N\right\}$, we employ a bi-directional GRU to incorporate the contextual information on argument level.
\begin{equation}
\setlength{\abovedisplayskip}{3pt}
\setlength{\belowdisplayskip}{3pt}
{h_i^{c}} = BiGRU(a_i,h_{i-1}^{c})
\end{equation}
Finally, we employ an average pooling over arguments to obtain the context representation $C$.

\subsection{Argument Matching and Scoring}
Once representations of the quotation and the reply are generated, three matching methods are applied to analyze relevance between the two arguments. We conduct element-wise product and element-wise difference to get the semantic features $f_{p}=R_q*R_r$ and $f_{d}=R_q-R_r$. Furthermore, to evaluate the relevance between each word in the reply and the discrete representation of the quotation, we propose the quotation-guided attention and obtain a new representation of the reply.

\noindent \textbf{Quotation-Guided Attention.} We conduct dot product between $R_{q}$ and each hidden state representation $h_j^r$ in the reply. Then, a softmax layer is used to obtain an attention distribution. 
\begin{equation}
\setlength{\abovedisplayskip}{3pt}
\setlength{\belowdisplayskip}{3pt}
v_j = softmax(R_{q} \cdot h_j^{r})
\end{equation}
Based on the attention probability $v_j$ of the $j$-th word in the reply, the new representation of the reply can then be constructed as follows:
\begin{equation}
\setlength{\abovedisplayskip}{3pt}
\setlength{\belowdisplayskip}{3pt}
f_r = \sum_{j} {{v_j} \cdot {h_j^{r}}}
\end{equation}
After obtaining the discrete representations, argumentative context representations and some semantic matching features $f_p$, $f_d$, $f_r$ of the quotation and the reply, we use two fully connected layers to obtain a higher-level representation $H$. Finally, the matching score $S$ is obtained by a linear transformation. 
\begin{equation}
\setlength{\abovedisplayskip}{3pt}
\setlength{\belowdisplayskip}{3pt}
f_m=[f_p;f_d;f_r]
\end{equation}
\begin{equation}
\setlength{\abovedisplayskip}{3pt}
\setlength{\belowdisplayskip}{3pt}
H = f(W_H[R_q;R_r;C_q;C_r;f_m]+b_H)
\end{equation}
\begin{equation}
\setlength{\abovedisplayskip}{3pt}
\setlength{\belowdisplayskip}{3pt}
S= W_{s}H+b_s
\end{equation}
where $W_H$ and $W_S$ stand for the weight matrices, while $b_H$ and $b_S$ are weight vectors.

\subsection{Joint Learning}
The proposed model contains three modules, i.e., the \emph{DVAE}, argumentative context modeling and argument matching, which are trained jointly. We define the loss function of the overall framework to combine the two effects.
\begin{equation}
\setlength{\abovedisplayskip}{3pt}
\setlength{\belowdisplayskip}{3pt}
L=L_{DVAE}+\lambda L_{m}
\end{equation}
where $\lambda$ is a hyper-parameter to balance the two loss terms.
The first loss term is defined on the \emph{DVAE} and cross entropy loss is defined as the reconstruction loss. We apply the regularization on \emph{KL} cost term to solve posterior collapse issue. Due to the space limitation, we leave out the derivation details and refer the readers to~\citeauthor{Zhao:18}~\shortcite{Zhao:18}.
\begin{equation}
\setlength{\abovedisplayskip}{3pt}
\setlength{\belowdisplayskip}{3pt}
L_{DVAE}=E_{q(z|x)}[\log p(x|z)]-KL(q(z|x)||p(z))
\end{equation}
The second loss term is defined on the argument matching. We formalize this issue as a ranking task and utilize hinge loss for training. 
\begin{equation}
\setlength{\abovedisplayskip}{3pt}
\setlength{\belowdisplayskip}{3pt}
L_{m}=\sum_{i=1}^{u} max(0,\gamma -S(q,r^{+})+S(q,r_i^{-}))
\label{equa_L}
\end{equation}
where $u$ is the number of negative replies in each instance. $\gamma$ is a margin parameter, $S(q,r^{+})$ is the matching score of the positive pair and $S(q,r_i^{-})$ is the matching score of the $i$-th negative pair.



\section{Experiment Setup}

\subsection{Training Details}
We use Glove~\cite{Pennington:14} word embeddings with dimension of 50.
The number of discrete latent variables \emph{M} is 5 and the number of categories for each latent variable is also 5. What’s more, the hidden units of GRU cell in encoder are 200 while that for the decoder is 400. We set batch size to 32, filter sizes to 5, filter numbers to 100, dropout with probability of 0.5, temperature $\tau$ to 1. The hyper-parameters in loss function are set as $\gamma$= 10 for max margin and $\lambda$= 1 for controlling the effects of discrete argument representation learning and argument matching. 

The proposed model is optimized by SGD and applied the strategy of learning rate decay with initial learning rate of 0.1. We evaluate our model on development set at every epoch to select the best model. During training, we run our model for 200 epochs with early-stop~\cite{Caruana:01}. 



\subsection{Comparison Models}
For baselines, we consider  simple models that rank argument pairs with cosine similarity measured with two types of word vectors: TF-IDF scores (henceforth \textsc{TF-IDF}) and the pre-trained word embeddings from word2vec corpus (henceforth \textsc{Word2Vec}). Also, we compare with the  neural models from related areas: \textsc{MaLSTM}~\cite{mueller:16}, the popular method for sentence-level semantic matching, and \textsc{CBCAWOF}~\cite{Ji:2018}, the state-of-the-art model to evaluate the persuasiveness of argumentative comments, which is tailored to fit our task.
In addition, we compare with some ablations to study the contribution from our components. 
Here we first consider \textsc{Match$_{rnn}$}, which uses BiGRU to learn argument representations and explore the match of arguments without modeling the context therein.
Then we compare with other ablations that adopt varying argument context modeling methods.
Here we consider BiGRU (henceforth \textsc{Match$_{rnn}$+C$_{b}$}), which focuses on words in argument context and ignores the argument interaction structure. We also consider a hierarchical neural network ablation (henceforth \textsc{Match$_{rnn}$+C$_{h}$}), which models argument interactions with BiGRU and the words therein with CNN. In addition, we compare with \textsc{Match$_{ae}$+C$_{h}$} and \textsc{Match$_{vae}$+C$_{h}$}, employing auto-encoder (AE) and variational AE (VAE), respectively, to take the duty of the \emph{DVAE} module of our full model.
\begin{table}[t!]\small
\setlength{\abovecaptionskip}{-0.02cm} 
\setlength{\belowcaptionskip}{-0.3cm} 
\begin{center}
\begin{tabular}{ |p{3.8cm} | p{1cm} | p{1cm} |}
\hline Models \centering & P@1  & MRR \\ \hline
\textbf{\underline{Cosine Similarity based}} & & \\
\textsc{TF-IDF} & 28.36*  & 51.66*  \\ 
\textsc{Word2vec} & 28.70*  & 52.03* \\ \hline\hline
\textbf{\underline{Neural-Network based}} & & \\
\textsc{MaLSTM}~\cite{mueller:16} & 31.26*   & 52.97*  \\
\textsc{CBCAWOF}~\cite{Ji:2018} & 56.04*   & 73.03*  \\ \hline\hline
\textbf{\underline{Ablation Study}} & & \\
\textsc{Match$_{rnn}$ }& 51.52*   & 70.57*  \\ 
\textsc{Match$_{rnn}$+C$_{b}$} & 55.98*  & 73.20* \\
\textsc{Match$_{rnn}$+C$_{h}$} &57.46*   & 73.72* \\
\textsc{Match$_{ae}$+C$_{h}$} & $\textrm{58.27}^\ddag$  &74.16*  \\ 
\textsc{Match$_{vae}$+C$_{h}$} & $\textrm{58.61}^\ddag$  & $\textrm{74.66}^\ddag$\\
\hline
\hline
\textbf{Our model} &\textbf{61.17}  &\textbf{76.16} \\ \hline
\end{tabular}
\end{center}
\caption{\label{font-table} The performances of different models on our dataset in terms of Mean Reciprocal Rank (\emph{MRR}) and Precision at 1 (denoted as \emph{P@1}). The proposed model significantly outperforms all the comparison methods marked with * or $^\ddag$ (*: \emph{p}$<$0.01; $^\ddag$: \emph{p}$<$0.05, Wilcoxon signed rank test). Best results are in \textbf{bold}. }
\label{tab:tb2}
\end{table}

\begin{figure}[t!]
\setlength{\abovecaptionskip}{-0.02cm} 
\setlength{\belowcaptionskip}{-0.3cm} 
\centering
  \includegraphics[width=0.9\columnwidth]{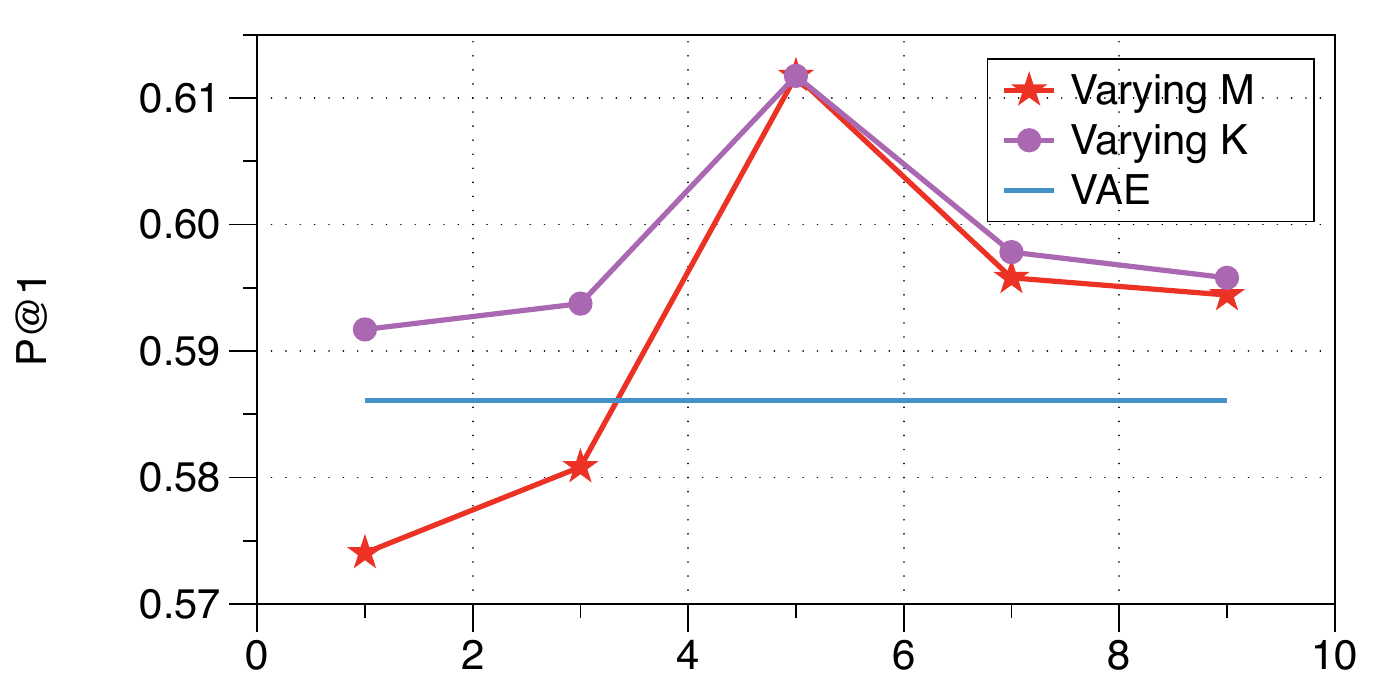}
    \caption{The impact of varying the number of discrete latent variables \emph{M} and categories for each latent variable \emph{K} on \emph{P@1}. We find that our model still outperforms \emph{VAE} which is the most competitive baseline. }
    \label{fig1}
\end{figure}
\section{Results and Discussions}
To evaluate the performance of different models, we first show the overall performance of different models for argument pair identification.
Then, we conduct three analyses including \emph{hyper-parameters sensitivity analysis}, \emph{discrete latent variables analysis} and \emph{error analysis} to study the impact of hyper-parameters on our model, explain why \emph{DVAE} performs well on interactive argument  pair identification and analyze the major causes of errors. Finally, we apply our model to a downstream task to further investigate the usefulness of discrete argument representations.

\subsection{Overall Performance Comparison}
The overall results of different models are shown in Table~\ref{tab:tb2}. Mean Reciprocal Rank (\emph{MRR}) and Precision at 1 (denoted as \emph{P@1}) are used for evaluation metrics. We have following findings.

\noindent - Our model significantly outperforms all comparison models in terms of both evaluation metrics. This proves the effectiveness of our model.

\noindent - Neural network models perform better than \textsc{TFIDF} and \textsc{Word2vec}. This observation shows the effectiveness of argument representation learning in neural networks.

\noindent - By modeling context representations, \textsc{Match$_{rnn}$+C$_{b}$} and \textsc{Match$_{rnn}$+C$_{h}$} significantly outperform \textsc{Match$_{rnn}$}. This proves that contextual information is helpful for identifying interactive argument pairs.

\noindent - Argumentative contexts often contain a list of arguments. In comparison of \textsc{Match$_{rnn}$+C$_{b}$} and \textsc{Match$_{rnn}$+C$_{h}$}, we find that \textsc{Match$_{rnn}$+C$_{h}$} achieve much better results than \textsc{Match$_{rnn}$+C$_{b}$}. This demonstrates the effectiveness of representing argumentative contexts on argument level instead of word level.

\noindent - By using autoencoders for argument representation learning, our model, \textsc{Match$_{vae}$+C$_{h}$} and \textsc{Match$_{ae}$+C$_{h}$} outperform \textsc{Match$_{rnn}$+C$_{h}$}. This indicates the effectiveness of argument representation learning.



\begin{figure*}[t!]
\setlength{\abovecaptionskip}{-0.02cm} 
\setlength{\belowcaptionskip}{-0.3cm} 
\subfigure{
\includegraphics[width=0.72\columnwidth,height=3.6cm] {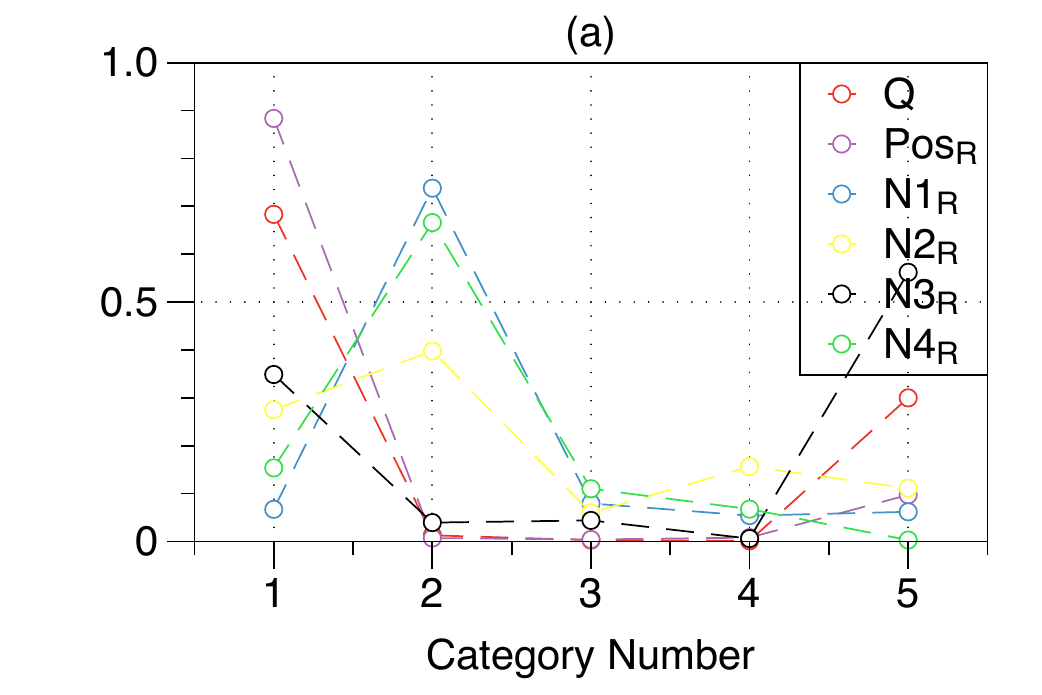}
}\hskip -24pt
\subfigure{
\includegraphics[width=0.72\columnwidth,height=3.6cm] {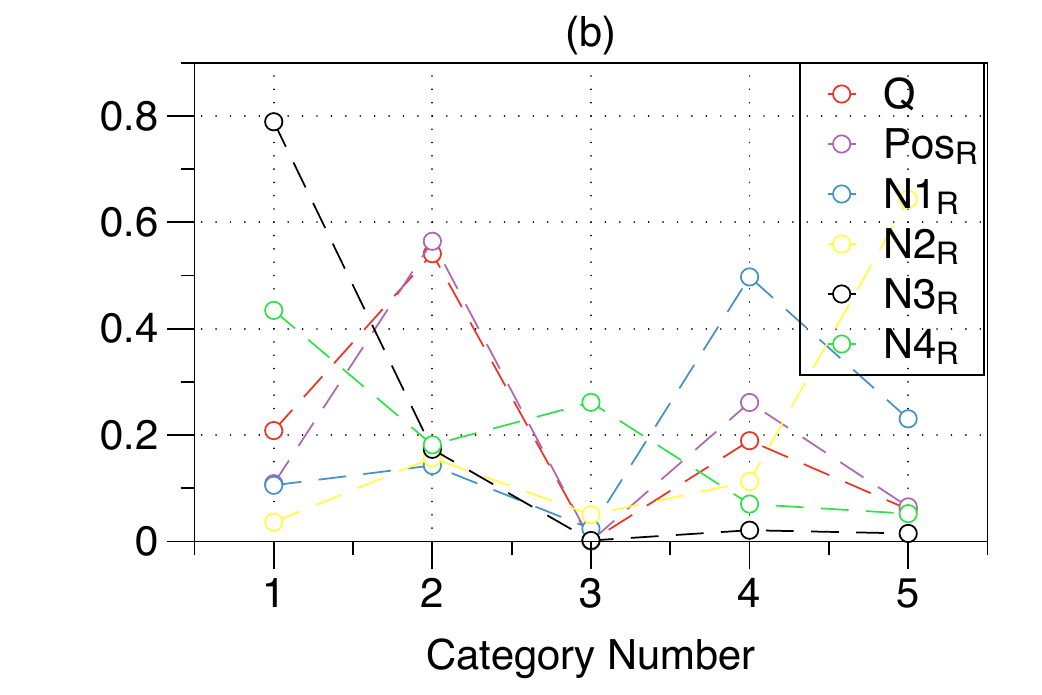}
}\hskip -24pt
\subfigure{
\includegraphics[width=0.72\columnwidth,height=3.6cm] {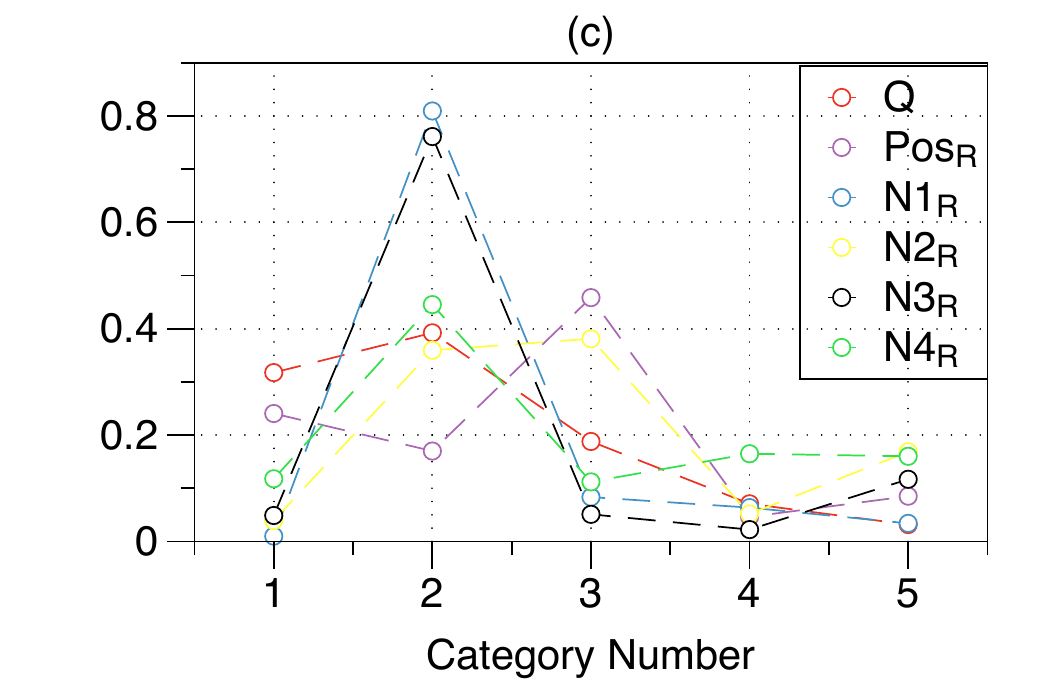}
}\hskip -24pt
\subfigure{
\includegraphics[width=0.72\columnwidth,height=3.6cm] {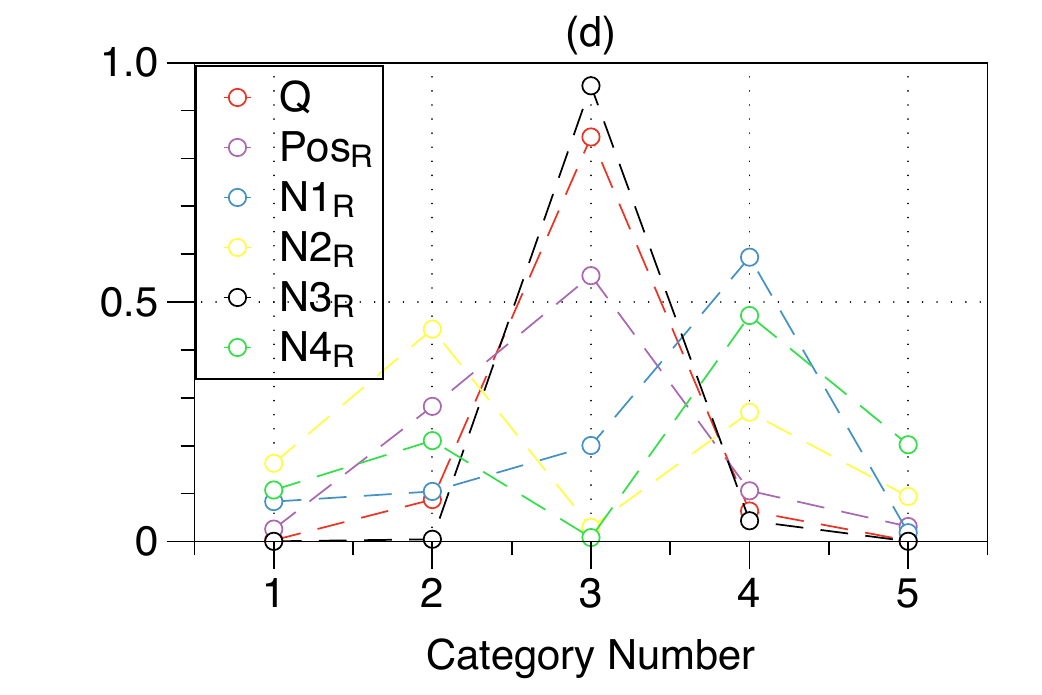}
} \hskip -24pt
\subfigure{
\includegraphics[width=0.72\columnwidth,height=3.6cm] {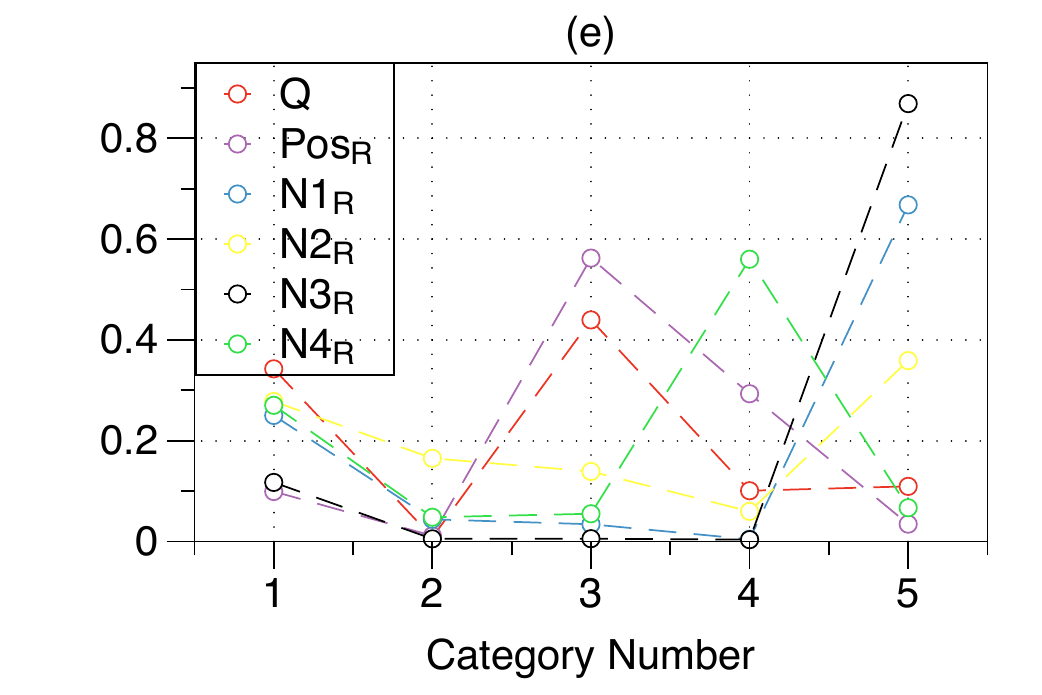}
}



\caption{Visualization of posterior distributions of discrete latent variables $z_1\sim z_5$ respectively. We find that the posterior distributions of $z_1\sim z_5$ of \textit{Positive reply} is more similar to those of \textit{Quotation} compared to other \textit{Negative replies}.}
\label{fig2}
\end{figure*}

\subsection{Hyper-Parameter Sensitivity Analysis}
We investigate the impact of two hyper-parameters on our model, namely the number of discrete latent variables \emph{M} and the number of categories for each latent variable \emph{K} in \emph{DVAE}. For studying the impact of \emph{M} and \emph{K}, we set them as 1, 3, 5, 7, 9 respectively while keep other hyper-parameters the same as our best model. We report \emph{P@1} of different settings. 

As shown in Figure~\ref{fig1}, we observe that curves obtained by changing the two parameters follow similar pattern. When the number increases, \emph{P@1} first gradually grows, reaching the highest at position 5 and drops gradually after that. When \emph{K} and \emph{M} are relatively high, say larger than 3, our model can always outperform \emph{VAE} which is the most competitive baseline, indicating the effectiveness of the discrete representation for interactive arguments identification.


\subsection{Discrete Latent Variables Analysis}
Here, we try to find out why \emph{DVAE} performs best on interactive argument pair identification. Given an argument, we set \emph{M}=5, \emph{K}=5 and learn the corresponding discrete code set $Z_{code}(1)\sim Z_{code}(5)$. We use the best model to select correct instances for argument matching in the dataset and cluster all quotations and corresponding replies according to the same discrete code set. We get 2,272 clusters, of which 119 clusters have more than 100 arguments and we find that arguments with the same discrete code set are semantically related. 

To show the reason why \emph{DVAE} performs well on our task more intuitively, we select a case from our dataset shown in Table~\ref{tab:tb11} and employ \emph{DVAE} to learn discrete representations for arguments to capture varying aspects $z_1\sim z_5$.  The posterior distributions of discrete latent variables $z_1\sim z_5$ for the quotation and replies are shown in Figure~\ref{fig2}. 

\begin{table}[t!]
\small
\setlength{\abovecaptionskip}{-0cm} 
\setlength{\belowcaptionskip}{-0.5cm} 
\begin{center}
\begin{tabular}{|p{0.45\textwidth}|}\hline 
\multicolumn{1}{|p{0.45 \textwidth}|}{\textbf{Quotation:}\ \ I bet that John Boehner would deal with congress as president more easily than Joe Biden due to his constant interaction with it.}\\
\hline 
\multicolumn{1}{|p{0.45 \textwidth}|}{\textbf{Positive reply:}\ \ Do you think that have anything to do with the fact that Boehner is a republican, and congress is controlled by republicans?}\\
\multicolumn{1}{|p{0.45 \textwidth}|}{\textbf{Negative reply 1:}\ \ I would propose that the title of vice president be kept, but to remove their right to succession for presidency.}\\
\multicolumn{1}{|p{0.45 \textwidth}|}{\textbf{Negative reply 2:}\ \ Does Biden have the same level of respect from foreign nations needed to guide the country?}\\
\multicolumn{1}{|p{0.45 \textwidth}|}{\textbf{Negative reply 3:}\ \ He did lose however, so perhaps people do put weight into the vp choice.}\\
\multicolumn{1}{|p{0.45 \textwidth}|}{\textbf{Negative reply 4:}\ \ I don't know why you think this can be ignored.}\\\hline 
\end{tabular}
\end{center}
\caption{\label{font-table} A case selected from our dataset.}
\label{tab:tb11}
\end{table}







As shown in Figure~\ref{fig2}, each subgraph shows the distribution of $z_i$ on \emph{K} categories of the quotation and corresponding replies. We can find that the posterior distributions of $z_1\sim z_5$ of \textit{Positive reply} are more similar to those of \textit{Quotation} compared to other \textit{Negative replies}. This finding proves that if the two arguments are more semantically related, their posterior distribution on each aspect $z_i$ should be more similar. This further interprets why \textit{Positive reply} has interactive relationship with \textit{Quotation} and why \emph{DVAE} performs well on interactive argument pair identification.

\subsection{Error Analysis}
Here, we inspect outputs of our model to identify major causes of errors. Here are two major issues.  

\noindent- The number of \emph{M} and \emph{K} may not cover the latent space of all arguments in the dataset.\ \ Natural language is complex and diverse. If the size of the latent space doesn't fully contain semantic information of the arguments, it will cause the failure of our model. Considering the number of aspects may varies for different topics, it is not perfect to use a universal setting of \emph{K} and \emph{M}. 

\noindent- Attention Error.\ \ In our model, we employ a quotation-guided attention to evaluate the relevance between each word in the reply and the discrete representation of the quotation. If the attention focuses on unimportant words, it causes errors. It might be useful to utilize discrete representation to further regulate the attention procedure. 

\section{Related Work}
In this section, we will introduce two major areas related to our work, which are dialogical argumentation and argument representation learning.
\subsection{Dialogical Argumentation}

Dialogical argumentation is a growing field in argumentation mining. Existing research covers discourse structure prediction~\cite{Wang:11,Liu:18}, dialog summarization~\cite{Hsueh:07}, etc. There are several attempts to address tasks related to analyzing the relationship between arguments~\cite{Wang:14,Persing:17} and evaluating the quality of persuasive arguments~\cite{HabernalIvan:16,WeiandLiu:16}.
However, there is limited research on the interactions between posts. In this work, we propose a novel task of identifying interactive argument pairs from argumentative posts to further understand the interactions between posts.

Our work is also related with some similar tasks, such as question answering and sentence alignment.
They focus on the design of attention mechanism to learn sentence representations~\cite{cui:2016,wang:2017gated} and their relations with others~\cite{chen:2016enhanced,wang:2017bilateral}.
Our task is inherently different from theirs because our target arguments naturally occur in the complex interaction context of dialogues, which requires additional efforts for understanding the discourse structure therein.




\subsection{Argument Representation Learning}
Argument representation learning for natural language has been studied widely in the past few years. Due to the availability of practically unlimited textual data, learning argument representations via unsupervised methods is an attractive proposition~\cite{Kiros:15,Bowman:16,Hill:16,Logeswaran:18}. 
Previous work focuses on learning continuous argument representations with no interpretability. In this work, we study the discrete argument representations, capturing varying aspects in argumentation languages.
\section{Conclusion and Future Work}
In this paper, we propose a novel task of interactive argument pair identification from two posts with opposite stances on a certain topic. We examine contexts of arguments and induce latent representations via discrete variational autoencoders. Experimental results on the dataset show that our model significantly outperforms the competitive baselines. Further analyses reveal why our model yields superior performance and prove the usefulness of discrete argument representations. 

The future work will be carried out in two directions. First, we will study the usage of our model for applying to other dialogical argumentation related tasks, such as debate summarization. Second, we will utilize neural topic model for learning discrete argument representations to further improve the interpretability of representations.

\bibliography{emnlp-ijcnlp-2019}

\begin{thebibliography}{32}
\expandafter\ifx\csname natexlab\endcsname\relax\def\natexlab#1{#1}\fi

\bibitem[{Asterhan and Schwarz(2007)}]{Asterhan:07}
Christa~SC Asterhan and Baruch~B Schwarz. 2007.
\newblock The effects of monological and dialogical argumentation on concept
  learning in evolutionary theory.
\newblock \emph{Journal of educational psychology}, 99(3):626.

\bibitem[{Bilu and Slonim(2016)}]{Bilu:16}
Yonatan Bilu and Noam Slonim. 2016.
\newblock Claim synthesis via predicate recycling.
\newblock In \emph{Proceedings of the 54th Annual Meeting of the Association
  for Computational Linguistics (Volume 2: Short Papers)}, volume~2, pages
  525--530.

\bibitem[{Botschen et~al.(2018)Botschen, Sorokin, and Gurevych}]{Botschen:18}
Teresa Botschen, Daniil Sorokin, and Iryna Gurevych. 2018.
\newblock \href {http://aclweb.org/anthology/W18-5211} {Frame- and entity-based
  knowledge for common-sense argumentative reasoning}.
\newblock In \emph{Proceedings of the 5th Workshop on Argument Mining}, pages
  90--96. Association for Computational Linguistics.

\bibitem[{Bowman et~al.(2016)Bowman, Vilnis, Vinyals, Dai, Jozefowicz, and
  Bengio}]{Bowman:16}
Samuel~R. Bowman, Luke Vilnis, Oriol Vinyals, Andrew Dai, Rafal Jozefowicz, and
  Samy Bengio. 2016.
\newblock \href {https://doi.org/10.18653/v1/K16-1002} {Generating sentences
  from a continuous space}.
\newblock In \emph{Proceedings of The 20th SIGNLL Conference on Computational
  Natural Language Learning}, pages 10--21. Association for Computational
  Linguistics.

\bibitem[{Caruana et~al.(2001)Caruana, Lawrence, and Giles}]{Caruana:01}
Rich Caruana, Steve Lawrence, and C~Lee Giles. 2001.
\newblock Overfitting in neural nets: Backpropagation, conjugate gradient, and
  early stopping.
\newblock In \emph{Advances in neural information processing systems}, pages
  402--408.

\bibitem[{Chen et~al.(2016)Chen, Zhu, Ling, Wei, Jiang, and
  Inkpen}]{chen:2016enhanced}
Qian Chen, Xiaodan Zhu, Zhenhua Ling, Si~Wei, Hui Jiang, and Diana Inkpen.
  2016.
\newblock Enhanced lstm for natural language inference.
\newblock \emph{arXiv preprint arXiv:1609.06038}.

\bibitem[{Cui et~al.(2016)Cui, Chen, Wei, Wang, Liu, and Hu}]{cui:2016}
Yiming Cui, Zhipeng Chen, Si~Wei, Shijin Wang, Ting Liu, and Guoping Hu. 2016.
\newblock Attention-over-attention neural networks for reading comprehension.
\newblock \emph{arXiv preprint arXiv:1607.04423}.

\bibitem[{Dong et~al.(2017)Dong, Zhang, and Yang}]{Dong:17}
Fei Dong, Yue Zhang, and Jie Yang. 2017.
\newblock Attention-based recurrent convolutional neural network for automatic
  essay scoring.
\newblock In \emph{Proceedings of the 21st Conference on Computational Natural
  Language Learning (CoNLL 2017)}, pages 153--162.

\bibitem[{Habernal and Gurevych(2016)}]{HabernalIvan:16}
Ivan Habernal and Iryna Gurevych. 2016.
\newblock Which argument is more convincing? analyzing and predicting
  convincingness of web arguments using bidirectional lstm.
\newblock In \emph{Proceedings of the 54th Annual Meeting of the Association
  for Computational Linguistics (Volume 1: Long Papers)}, volume~1, pages
  1589--1599.

\bibitem[{Hill et~al.(2016)Hill, Cho, and Korhonen}]{Hill:16}
Felix Hill, Kyunghyun Cho, and Anna Korhonen. 2016.
\newblock \href {https://doi.org/10.18653/v1/N16-1162} {Learning distributed
  representations of sentences from unlabelled data}.
\newblock In \emph{Proceedings of the 2016 Conference of the North American
  Chapter of the Association for Computational Linguistics: Human Language
  Technologies}, pages 1367--1377. Association for Computational Linguistics.

\bibitem[{Hsueh and Moore(2007)}]{Hsueh:07}
Pei-Yun Hsueh and Johanna~D Moore. 2007.
\newblock Automatic decision detection in meeting speech.
\newblock In \emph{International Workshop on Machine Learning for Multimodal
  Interaction}, pages 168--179. Springer.

\bibitem[{Hunter(2013)}]{Hunter:13}
Anthony Hunter. 2013.
\newblock Analysis of dialogical argumentation via finite state machines.
\newblock In \emph{International Conference on Scalable Uncertainty
  Management}, pages 1--14. Springer.

\bibitem[{Ji et~al.(2018)Ji, Wei, Hu, Liu, Zhang, and Huang}]{Ji:2018}
Lu~Ji, Zhongyu Wei, Xiangkun Hu, Yang Liu, Qi~Zhang, and Xuanjing Huang. 2018.
\newblock Incorporating argument-level interactions for persuasion comments
  evaluation using co-attention model.
\newblock In \emph{Proceedings of the 27th International Conference on
  Computational Linguistics}, pages 3703--3714.

\bibitem[{Kiros et~al.(2015)Kiros, Zhu, Salakhutdinov, Zemel, Urtasun,
  Torralba, and Fidler}]{Kiros:15}
Ryan Kiros, Yukun Zhu, Ruslan~R Salakhutdinov, Richard Zemel, Raquel Urtasun,
  Antonio Torralba, and Sanja Fidler. 2015.
\newblock Skip-thought vectors.
\newblock In \emph{Advances in neural information processing systems}, pages
  3294--3302.

\bibitem[{Liu et~al.(2018)Liu, Cohen, and Lapata}]{Liu:18}
Jiangming Liu, Shay~B. Cohen, and Mirella Lapata. 2018.
\newblock \href {http://aclweb.org/anthology/P18-1040} {Discourse
  representation structure parsing}.
\newblock In \emph{Proceedings of the 56th Annual Meeting of the Association
  for Computational Linguistics (Volume 1: Long Papers)}, pages 429--439.
  Association for Computational Linguistics.

\bibitem[{Logeswaran and Lee(2018)}]{Logeswaran:18}
Lajanugen Logeswaran and Honglak Lee. 2018.
\newblock An efficient framework for learning sentence representations.
\newblock In \emph{ICLR}.

\bibitem[{Lu et~al.(2017)Lu, Kannan, Yang, Parikh, and Batra}]{Lu:17}
Jiasen Lu, Anitha Kannan, Jianwei Yang, Devi Parikh, and Dhruv Batra. 2017.
\newblock Best of both worlds: Transferring knowledge from discriminative
  learning to a generative visual dialog model.
\newblock In \emph{Advances in Neural Information Processing Systems}, pages
  314--324.

\bibitem[{Mueller and Thyagarajan(2016)}]{mueller:16}
Jonas Mueller and Aditya Thyagarajan. 2016.
\newblock Siamese recurrent architectures for learning sentence similarity.
\newblock In \emph{Thirtieth AAAI Conference on Artificial Intelligence}.

\bibitem[{Pennington et~al.(2014)Pennington, Socher, and
  Manning}]{Pennington:14}
Jeffrey Pennington, Richard Socher, and Christopher Manning. 2014.
\newblock Glove: Global vectors for word representation.
\newblock In \emph{Proceedings of the 2014 conference on empirical methods in
  natural language processing (EMNLP)}, pages 1532--1543.

\bibitem[{Persing and Ng(2017)}]{Persing:17}
Isaac Persing and Vincent Ng. 2017.
\newblock Why can’t you convince me? modeling weaknesses in unpersuasive
  arguments.
\newblock In \emph{Proceedings of the 26th International Joint Conference on
  Artificial Intelligence}, pages 4082--4088. AAAI Press.

\bibitem[{Rolfe(2017)}]{Rolfe:17}
Jason~Tyler Rolfe. 2017.
\newblock Discrete variational autoencoders.
\newblock In \emph{ICLR}.

\bibitem[{Sanchan et~al.(2017)Sanchan, Aker, and Bontcheva}]{Sanchan:17}
Nattapong Sanchan, Ahmet Aker, and Kalina Bontcheva. 2017.
\newblock \href {https://doi.org/10.26615/978-954-452-038-0_003} {Automatic
  summarization of online debates}.
\newblock In \emph{Proceedings of the 1st Workshop on Natural Language
  Processing and Information Retrieval associated with RANLP 2017}, pages
  19--27. INCOMA Inc.

\bibitem[{Stab and Gurevych(2014)}]{Stab:14}
Christian Stab and Iryna Gurevych. 2014.
\newblock Identifying argumentative discourse structures in persuasive essays.
\newblock In \emph{Proceedings of the 2014 Conference on Empirical Methods in
  Natural Language Processing (EMNLP)}, pages 46--56.

\bibitem[{Taghipour and Ng(2016)}]{Taghipour:16}
Kaveh Taghipour and Hwee~Tou Ng. 2016.
\newblock A neural approach to automated essay scoring.
\newblock In \emph{Proceedings of the 2016 Conference on Empirical Methods in
  Natural Language Processing}, pages 1882--1891.

\bibitem[{Tan et~al.(2016)Tan, Niculae, Danescu-Niculescu-Mizil, and
  Lee}]{Tan:16}
Chenhao Tan, Vlad Niculae, Cristian Danescu-Niculescu-Mizil, and Lillian Lee.
  2016.
\newblock Winning arguments: Interaction dynamics and persuasion strategies in
  good-faith online discussions.
\newblock In \emph{Proceedings of the 25th international conference on world
  wide web}, pages 613--624. International World Wide Web Conferences Steering
  Committee.

\bibitem[{Wang et~al.(2011)Wang, Lui, Kim, Nivre, and Baldwin}]{Wang:11}
Li~Wang, Marco Lui, Su~Nam Kim, Joakim Nivre, and Timothy Baldwin. 2011.
\newblock \href {http://aclweb.org/anthology/D11-1002} {Predicting thread
  discourse structure over technical web forums}.
\newblock In \emph{Proceedings of the 2011 Conference on Empirical Methods in
  Natural Language Processing}, pages 13--25. Association for Computational
  Linguistics.

\bibitem[{Wang and Cardie(2014)}]{Wang:14}
Lu~Wang and Claire Cardie. 2014.
\newblock \href {https://doi.org/10.3115/v1/P14-2113} {A piece of my mind: A
  sentiment analysis approach for online dispute detection}.
\newblock In \emph{Proceedings of the 52nd Annual Meeting of the Association
  for Computational Linguistics (Volume 2: Short Papers)}, pages 693--699.
  Association for Computational Linguistics.

\bibitem[{Wang et~al.(2018)Wang, Wang, and Zhang}]{Wang:18}
Nan Wang, Jin Wang, and Xuejie Zhang. 2018.
\newblock \href {https://doi.org/10.18653/v1/S18-1073} {Ynu-hpcc at
  semeval-2018 task 2: Multi-ensemble bi-gru model with attention mechanism for
  multilingual emoji prediction}.
\newblock In \emph{Proceedings of The 12th International Workshop on Semantic
  Evaluation}, pages 459--465. Association for Computational Linguistics.

\bibitem[{Wang et~al.(2017{\natexlab{a}})Wang, Yang, Wei, Chang, and
  Zhou}]{wang:2017gated}
Wenhui Wang, Nan Yang, Furu Wei, Baobao Chang, and Ming Zhou.
  2017{\natexlab{a}}.
\newblock Gated self-matching networks for reading comprehension and question
  answering.
\newblock In \emph{Proceedings of the 55th Annual Meeting of the Association
  for Computational Linguistics (Volume 1: Long Papers)}, pages 189--198.

\bibitem[{Wang et~al.(2017{\natexlab{b}})Wang, Hamza, and
  Florian}]{wang:2017bilateral}
Zhiguo Wang, Wael Hamza, and Radu Florian. 2017{\natexlab{b}}.
\newblock Bilateral multi-perspective matching for natural language sentences.
\newblock \emph{arXiv preprint arXiv:1702.03814}.

\bibitem[{Wei et~al.(2016)Wei, Liu, and Li}]{WeiandLiu:16}
Zhongyu Wei, Yang Liu, and Yi~Li. 2016.
\newblock Is this post persuasive? ranking argumentative comments in online
  forum.
\newblock In \emph{Proceedings of the 54th Annual Meeting of the Association
  for Computational Linguistics (Volume 2: Short Papers)}, volume~2, pages
  195--200.

\bibitem[{Zhao et~al.(2018)Zhao, Lee, and Eskenazi}]{Zhao:18}
Tiancheng Zhao, Kyusong Lee, and Maxine Eskenazi. 2018.
\newblock \href {http://aclweb.org/anthology/P18-1101} {Unsupervised discrete
  sentence representation learning for interpretable neural dialog generation}.
\newblock In \emph{Proceedings of the 56th Annual Meeting of the Association
  for Computational Linguistics (Volume 1: Long Papers)}, pages 1098--1107.
  Association for Computational Linguistics.

\end{thebibliography}
\bibliographystyle{acl_natbib}

\end{document}